\documentclass[10pt,fleqn,a4paper,twoside]{article}
\usepackage{eurodiname2026}
\usepackage{amsmath, bm} % assumes amsmath package installed
\usepackage{amsfonts}
\usepackage{amssymb}  % assumes amsmath package installed
\usepackage{hyperref}
\usepackage{caption}
\usepackage{subcaption}
\usepackage{float}
\def\shortauthor{A. Alwala, G. S. Lima and W. M. Bessa}
\def\shorttitle{Machine Learning-Based Feedback Linearization Control of Quadrotor Subject to Unmodeled Dynamics}

\begin{document}
\fphead
\hspace*{-2.5mm}\begin{tabular}{||p{\textwidth}}
\begin{center}
\vspace{-4mm}
\title{EURODINAME-2026-13273\\
MACHINE LEARNING-BASED FEEDBACK LINEARIZATION CONTROL OF QUADROTOR SUBJECT TO UNMODELED DYNAMICS} %(XXXX is the manuscript number. It will be available after the extended abstract submission and must be placed for the final paper submission.)
\end{center}
\authors{Amos Alwala} \\
\authors{Gabriel da Silva Lima} \\
\authors{Wallace Moreira Bessa} \\
\institution{\href{https://sites.utu.fi/smartsystems/}{Smart Systems}, Department of Mechanical and Materials Engineering, Faculty of Technology, University of Turku, 20520 Turku, Finland.} \\ %(If all authors are from the same institution, the "Institution and address" must be placed only once.)
\institution{amos.alwala@utu.fi, gdasil@utu.fi, wallace.moreirabessa@utu.fi} \\
\\
\\
\abstract{\textbf{Abstract.} The control of agile quadrotors in dynamic and uncertain environments remains an open area of investigation to this day, particularly when the complete system dynamics are partially known or highly nonlinear. This work introduces a novel machine learning-based feedback-linearization control framework that employs a Gaussian Radial Basis Function (RBF) neural network (NN) to model and compensate for unmodeled dynamics in real time. The proposed controller leverages the universal approximation capability of RBF networks to model nonlinearities and uncertainties. An online adaptation of the RBF NN updates the network's weights without prior training. The control law is derived using the Lyapunov stability theory, herein guaranteeing closed-loop stability and providing theoretical guarantee of asymptotic convergence of a trajectory tracking task. Gazebo simulation and real flight experiments are conducted using the Bitcraze's Crazyflie 2.1 quadrotor subject to unmodeled air drag, actuator dynamics, and external disturbance. Despite incomplete knowledge of prior dynamics and presence of external disturbance such as air drag and drift in state estimation, the proposed controller improves trajectory tracking with rapid convergence and reduction of position-norm and yaw orientation RMSE by more than $7.13\%$ and $49.27\%$ respectively compared to baseline feedback linearization controller.}\\
\\
\keywords{\textbf{Keywords:} Intelligent Control, Adaptive Control, Feedback Linearization, Radial Basis Function Networks}\\
\end{tabular}

\section{INTRODUCTION}

Aeriel robotics has seen widespread integration across diverse domains, including environmental monitoring, mapping, surveillance, and search-and-rescue operations. Among these, quadrotor Unmanned Arieal Vehicles (UAVs) have emerged as a more reliable platform owing to their agility, and capability to navigate complex, dynamic environments like forests and buildings characterized by confined spaces and dense obstacles. From a control perspective, quadrotor UAVs are categorized in the class of underactuated systems that possesses six degrees of freedom (6-DOF) governed by only four control inputs. The inherent underactuation, combined with strong non-linear coupling between translational and rotational dynamics necessitates high-fidelity modeling and sophisticated control strategies. Control approaches in literature include Proportional Integral Derivative (PID) and its variants (especially P and PD), Linear Quadratic Regulator (LQR), Model Predictive Control (MPC), Sliding Mode Control (SMC), Geometric Control, Feedback Linearization (FBL) Control, Convolutional Neural Network, and Reinforcement Learning. 

Classical control approaches such as PID and LQR have often been adopted for their simplicity; however, these rely on linearization around a specific operating point, often the hover state. This limits their stability and performance guarantee when the system deviates from the equilibrium manifold. To address these limitations, Sliding Mode Control has been extensively utilized due to its fundamental robustness against model inaccuracies and external disturbance. Significant contributions include PD-based sliding surface~\citep{Herrera}, rate-bounded PID integration~\citep{Rong}, and hierarchical sliding manifolds for decoupled subsystems~\citep{ZHENG20141350}. Although SMC provides superior disturbance rejection demonstrated by works such as; a robust differentiator for noise handling~\citep{Zhao}, and Higher-Order SMC for complex payloads~\citep{Chandra}, it is often affected by the chattering effect. These finite-frequency oscillations can excite unmodeled high-frequency dynamics and cause mechanical fatigue, a challenge that conventional SMC formulations struggle to mitigate.
%\citep{ZHENG20141350} and  proposed a fully actuated subsystem sliding manifold considered combining position and velocity tracking error of a single state variable whereas the underactuated subsystem, sliding manifold was a linear combination of position and velocity tracking errors of two state variables.
%In \citep{Zhao} an active disturbance rejection switching control based on an exact robust differentiator was proposed to handle white-noise disturbance above 0.1 dB intensity.
%In \citep{Chandra} proposed a higher order SMC for quadrotor control with dynamics that includes suspended load dynamics. a super twisting control input was derived to suppress load swings. These oscillations present an effect known as chattering (unwanted finite frequency oscillations) to the control system, which the conventional SMC cannot handle.

Model Predictive Control offers a sophisticated alternative by explicitly handling state and input constraints~\citep{pereira_nonlinear_2021}. However, many MPC implementations treat translation and rotational dynamics as decoupled entities, thereby neglecting the cross-axis dependencies that dictate high-performance maneuvers. In~\citep{Gang_cao_2016}, a hierarchical control was proposed with different MPC formulations for the translation and rotational subsystems. While approaches employing dual-quaternion manifolds~\citep{RecaldeLuisF} provide a globally non-singular framework for coupled dynamics representation, the primary bottleneck for MPC remains its high computational complexity, which often restricts its real-time implementation.

Recently, Reinforcement Learning and Neural-Network-based controllers have demonstrated remarkable success, particularly in high speed drone racing~\citep{song}. By leveraging domain randomization, these agents develop policies that are highly resilient to system uncertainties. Nevertheless, these rely on extensive offline training in simulation, and the use of such black-box models makes it difficult to ensure stability during online tracking of continuously differentiable 3D trajectories.
% These require training of an agent to learn a control policy in simulation via offline optimization by maximize rewards by trial and error while exploring an environment and taking feedback in the form of rewards and penalities for control actions.

%Classical model-based controllers often require accurate system identification, and their performance significantly degrades under modeling errors, unmodeled dynamics, or external disturbances. By integrating data-driven function approximation with traditional control theory, this research establishes a generalizable framework for intelligent flight control, advancing the development of adaptive and autonomous aerial robotic systems capable of reliable operation in unstructured environments.

In this paper, a Lyapunov-based nonlinear controller and Adaptive Neural Networks (ANN) is presented. This intelligent control (IC) scheme builds upon methodologies of ~\citep{Dos_Santos, Bessa}, and is derived based on Lagrangian dynamics of the quadrotor. First, the closed-loop stability of the feedback linearization controller is derived using the Lyapunov-based approach, which provides theoretical guarantees of asymptotic convergence in a trajectory tracking task. Then a Gaussian RBF network is chosen as the architecture for the ANN because of its universal approximation property to model nonlinearities. An online adaptation scheme updates the weights of the RBF function. This is done as an alternative to relying on rigid, predefined disturbance observers. By integrating an online machine learning scheme to adaptively model environmental disturbances in real-time, we allow for more precise compensation as the vehicle interacts with its surroundings. 
Furthermore, to address the inherent coupling, this work adopts a geometric control formulation on the Special Orthogonal Group $SO(3)$. Similarly to the methodology in~\citep{Antal}, we achieve coupling by deriving the desired attitude directly from the translational force vector. This approach ensures that the orientation and angular velocity error terms evolve naturally on the tangent space of the configuration manifold, herein maintaining global stability without the singularities associated with Euler angles.
% TO_CONSIDER: projection algorithm to further ensure stability

\begin{figure}[h!]
\centering
\includegraphics[width=140mm]{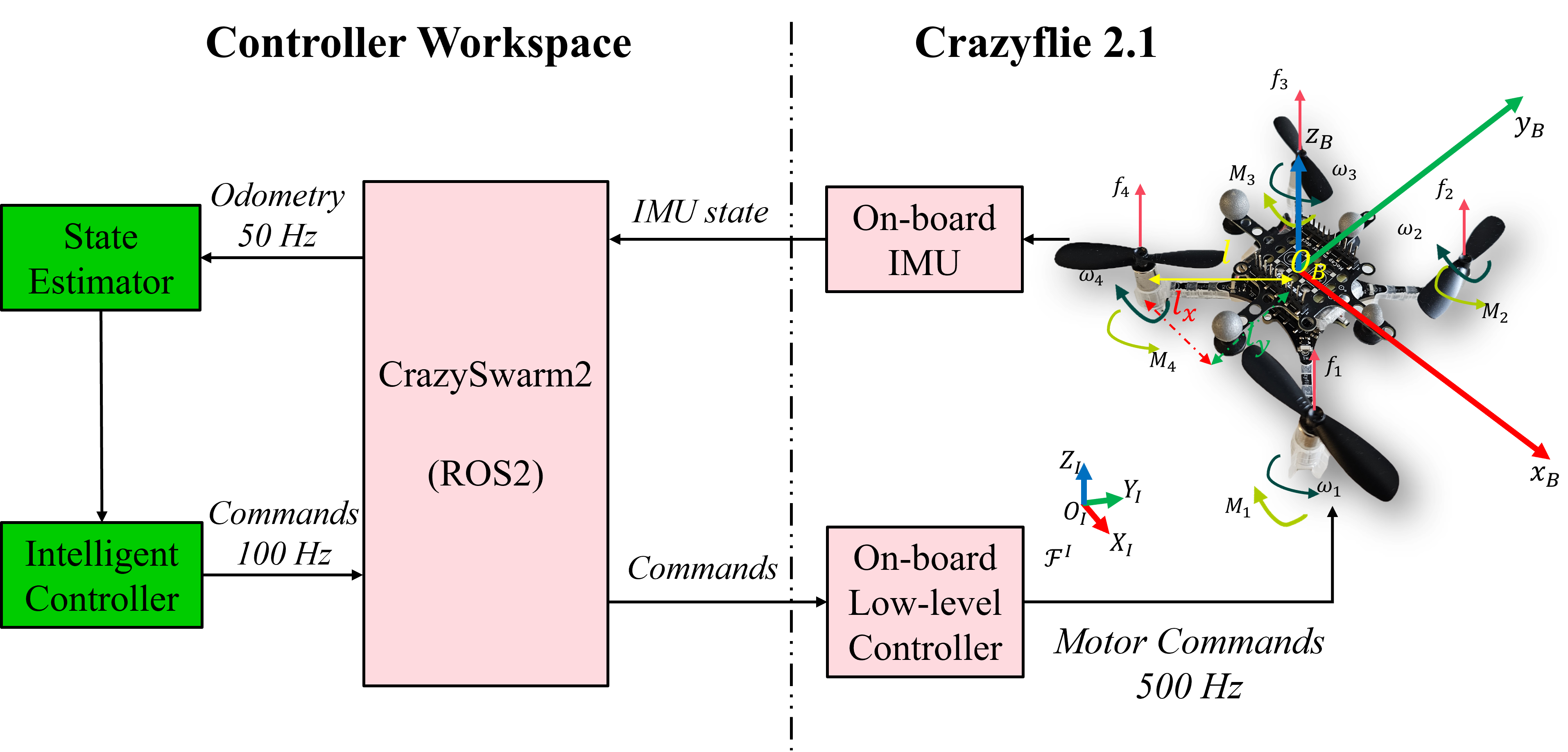}
\caption{System pipeline and Quadrotor layout showing the inertial and body frames.}
\label{quadcopter_pipeline}
\end{figure}

The system pipeline is presented in figure~\ref{quadcopter_pipeline}. We utilize the \href{https://www.bitcraze.io}{Bitcraze's Crazyflie 2.1} quadrotor, a versatile open source development kit that weighs about $29g$ and is equipped with a low-latency long-range radio. The quadrotor connects to a ground station computer using the Crazyradio PA and \href{https://github.com/IMRCLab/crazyswarm2}{CrazySwarm2}\citep{crazyswarm} which provide the communication link and ROS2 integration with the Crazyflie. The presented approach is first evaluated using numeric simulation and then implemented using Robot Operating System (ROS)-2 Humble with \href{https://github.com/knmcguire/ros_gz_crazyflie}{Gazebo Harmonic simulation} and later on real flights. Despite incomplete knowledge of prior dynamics and the presence of external disturbances, the proposed approach achieved precise trajectory tracking with a reduced error of up to $7.13\%$ in translation and $49.27\%$ in yaw on a real flight trajectory tracking task over the baseline. We open source the code release used in this paper at the \href{https://gitlab.utu.fi/smartsystems/research/releases/crazyflie_ros2_int_controller.git} {Smart System GitLab} page: https://gitlab.utu.fi/smartsystems/research/releases/crazyflie\_ros2\_int\_controller.git.

The rest of the paper details the quadrotor kinematics and dynamics in Section~\ref{kinematics and dynamics}. The formulation of the intelligent controller and the derivation of a Lyapunov-based control law is presented in  Section~\ref{control synthesis}. Simulation and real flight experiments conducted with the quadrotor are discussed in Section~\ref{experimentation}. Conclusions and future work are presented in Section~\ref{conclusion}.

\section{DRONE KINEMATICS AND DYNAMICS} \label{kinematics and dynamics}

The quadrotor is modeled as a rigid body in 3D space, that is, the shape and dimensions do not change over time. This assumption simplifies the analysis of the kinematics and dynamics, allowing for the application of rigid body mechanics to describe quadrotor motion. The mathematical models that capture its translational and rotational movements can then be derived, which are crucial for developing effective control strategies.

\subsection{Kinematics}

Figure~\ref{quadcopter_pipeline} indicates the representation of the quadrotor layout. Two coordinate systems are defined to describe motion in 3D space, that is, an inertial frame $\mathcal{F}^I$ fixed with respect to the world, and the body frame $\mathcal{F}^B$ fixed to the quadrotor centre of mass (COM). %"dfn other drone symbols". 
The COM is at a point $(x,y,z)$ from the origin of the inertial frame and is oriented at angles $\phi$ about the $x$-axis, $\theta$ about the $y$-axis, and $\psi$ about the $z$-axis of the body frame, which are Roll, Pitch and Yaw, respectively.
The transformation between the defined coordinate frames is described by a combination of translation and rotation. The rotation matrix follows the ZYX Euler convention and is defined as:
\begin{subequations} \label{eg:rot_matrix}
\begin{align}
\bm{R} &= Rot(z,\psi)Rot(y, \theta)Rot(x,\phi)  \label{eg:rot_matrixa}\\
\bm{R} &= \begin{bmatrix}c\theta c\psi & s\phi s\theta c\psi - c\phi s\psi & c\phi s\theta c\psi + s\phi s\psi \\ s\theta s\psi & s\phi s\theta s\psi - c\phi c\psi & c\phi s\theta s\psi - s\phi c\psi \\-s\theta & s\phi c\theta & c\phi c\theta \end{bmatrix} \label{eg:rot_matrixb}
\end{align}
\end{subequations}
where, $c\cdot$ and $s\cdot$ represent the cosine and sine of an angle. The rotation matrix $\bm{R} \in SO(3)$ belongs to the special orthogonal group of matrices for which the following properties hold:
\begin{equation}
SO(3) = \{\bm{R} \in \mathbb{R}^{3\times3} | \bm{R}\bm{R}^\top = \bm{R}^\top \bm{R} = \bm{I}, \det(\bm{R}) = 1 \}
\label{so3}
\end{equation}
where, $\bm{I}$ is a $3\times3$ identity matrix, and $\det(\cdot)$ denotes the determinant. The transformation matrix obtained as a combination of the 3D translation and rotation forms the Special Euclidean groups SE(3). 

Following the properties of the rotation matrix described in Eq.~\eqref{so3}, the relation between inertial rates and body rates can be derived from:
\begin{equation}
    \bm{R}\bm{R}^\top = \bm{R}^\top \bm{R} = \bm{I}
\end{equation}
whose first derivative is:
\begin{equation}
    \dot{\bm{R}}\bm{R}^\top + \bm{R}\dot{\bm{R}}^\top = \dot{\bm{R}}^\top \bm{R} + \bm{R}^\top \dot{\bm{R}} = 0
\end{equation}
where, $\dot{\bm{R}}\bm{R}^\top$ and $\bm{R}\dot{\bm{R}}^\top$ are skew-symmetric. For rigid body transformation, $\dot{\bm{R}} = \bm{R} \hat{\bm{\omega}}_b = \hat{\bm{\omega}}_i\bm{R}$, therefore:
\begin{subequations} \label{eq:omega}
\begin{align}
    \hat{\bm{\omega}}_b &= \bm{R}^\top \dot{\bm{R}} \label{eq:omegab} \\
    \hat{\bm{\omega}}_i &= \dot{\bm{R}}\bm{R}^\top \label{eq:omegai}
\end{align}
\end{subequations}
where $(\hat{.})$ is the hat operator which denotes a skew map. Eq.~\eqref{eq:omegab} encodes the angular velocity in the body frame. It can therefore be found that:
\begin{equation}
    \bm{\omega}_b = \begin{bmatrix}1 & 0 & -s\theta \\ 0 & c\phi & s\phi c\theta \\ 0 & -s\phi & c\phi c\theta\end{bmatrix} \bm{\omega}_i
\end{equation}
where, $\bm{\omega}_b$ is the body rate and $\bm{\omega}_i = [\dot{\phi}, \dot{\theta}, \dot{\psi}]$ is the inertial rate.

\subsection{Dynamics}

The quadrotor dynamics model is derived using the Lagrangian, $\frac{d}{dt}(\frac{\partial \mathcal{L}}{\partial \bm{\dot{q}}}) - \frac{\partial \mathcal{L}}{\partial \bm{q}} = \bm{F}$, where $\mathcal{L} = k_e - p_e$ is the Lagrangian, in which $k_e$ and $p_e$ are the kinetic and potential energies, respectively, and are derived as:
\begin{subequations}
\begin{align}
    k_e &= \frac{1}{2}m(\dot{x}^2 + \dot{y}^2 + \dot{z}^2) + \frac{1}{2} \bm{\omega}_b^\top \bm{J}\bm{\omega}_b \\
    p_e &= mgz
\end{align}
\end{subequations}
where $\bm{J} = \text{diag}(I_{xx}, I_{yy}, I_{zz}) $ is the quadrotor inertia matrix and $I_{xx}, I_{yy}, I_{zz}$ are the inertia of the quadrotor in $x$, $y$, and $z$ respectively.

By computing the Lagrangian in generalized coordinates $\bm{q} = [x, y, z, \phi, \theta, \psi]\top$, the dynamics of the quadrotor can be represented as:
\begin{equation}\label{eq:model}
    \bm{M}(\bm{q})\bm{\ddot{q}} + \bm{C}(\bm{q, \dot{q}})\bm{\dot{q}} + \bm{G}(\bm{q}) + \bm{\tau_d} = \bm{F}
\end{equation}
where, $\bm{M}(\bm{q}) \in \mathbb{R}^{6\times6}$ is the inertia matrix, $\bm{C}(\bm{q,\dot{q}}) \in \mathbb{R}^{6\times6}$ is the centripetal and coriolisis matrix whose entries are defined in the Appendix (Section~\ref{appendix}), $\bm{G}(\bm{q}) \in \mathbb{R}^{6\times1}$ is the gravity term, and $\bm{F} \in \mathbb{R}^{6\times1}$ is the generalized force vector that includes the translational force vector and torques in $RPY$. $\bm{\tau_d} \in \mathbb{R}^{6\times1}$ is introduced to represent all unmodeled dynamics including air drag, actuator dynamics, and other external disturbances. It can be observed that the quadrotor dynamics in Eq.~\eqref{eq:model} can be decoupled into a cascaded control system with translation and rotational dynamics as:
\begin{subequations}
\begin{align}
    \begin{bmatrix}m &0 &0 \\0 &m &0 \\0 &0 &m\end{bmatrix}\begin{bmatrix}\ddot{x} \\ \ddot{y} \\ \ddot{z}\end{bmatrix} + \begin{bmatrix} 0 \\ 0 \\ mg \end{bmatrix} + \begin{bmatrix} \tau_{d_x} \\ \tau_{d_y} \\ \tau_{d_z} \end{bmatrix} &=  \begin{bmatrix} F_x \\ F_y \\ F_z \end{bmatrix} \\
    \begin{bmatrix}m_{3,3} &0 &m_{3,5} \\0 &m_{4,4} &m_{4,5} \\m_{5,3} &m_{5,4} &m_{5,5}\end{bmatrix}\begin{bmatrix}\ddot{\phi} \\ \ddot{\theta} \\ \ddot{\psi}\end{bmatrix} + \begin{bmatrix}0 &c_{3,4} &c_{3,5} \\c_{4,3} &c_{4,4} &c_{4,5} \\c_{5,3} &c_{5,4} &c_{5,5}\end{bmatrix}\begin{bmatrix}\dot{\phi} \\ \dot{\theta} \\ \dot{\psi}\end{bmatrix} + \begin{bmatrix} \tau_{d_\phi} \\ \tau_{d_\theta} \\ \tau_{d_\psi} \end{bmatrix} &=  \begin{bmatrix} \tau_x \\ \tau_y \\ \tau_z \end{bmatrix}
\end{align}
\end{subequations}
And the thrust force, $T$ can be computed as:
\begin{equation}
    T = \bm{F}_{[:3]}^\top \bm{R}\bm{e}_3
\end{equation}
where $\bm{e}_3 = [0, 0, 1]^\top$ is a unit vector.

\section{INTELLIGENT CONTROL}  \label{control synthesis}

Intelligent control schemes possess the ability to predict the outcome of their control action, and this is achieved through learned interactions with their environment. They have found application in uncertain nonlinear systems for tasks such as  trajectory tracking~\citep{Da_Silva_Lima} and UAV formation ~\citep{Bui_Phung_2024}.

For the model presented in Eq.~\eqref{eq:model}, the feedback linearization control law~\citep{slotine1991applied} is proposed below:
\begin{equation}\label{eq:int_ctl_law}
    \bm{F} = \bm{M}(\bm{\ddot{q}_r} - 2\bm{\Lambda\dot{\tilde{q}}} - \bm{\Lambda^2\tilde{q}} - \bm{\hat{d}}) + \bm{C}\bm{\dot{q}} + \bm{G}
\end{equation}
where $\bm{\ddot{q}}_r \in \mathbb{R}^{6\times1}$ is the derivative of the reference velocity, $\bm{\tilde{q}}=\bm{q}-\bm{q}_r$ is the position and orientation tracking error, $\bm{\Lambda} \in \mathbb{R}^{6\times6}$ is a diagonal matrix with positive entries $\lambda_i$ the control gains, and $\bm{\hat{d}}$ is the estimate of $\bm{\tau}_d$ (from here on represented as $\bm{d}$).
\begin{figure}[h!]
\centering
\includegraphics[width=80mm]{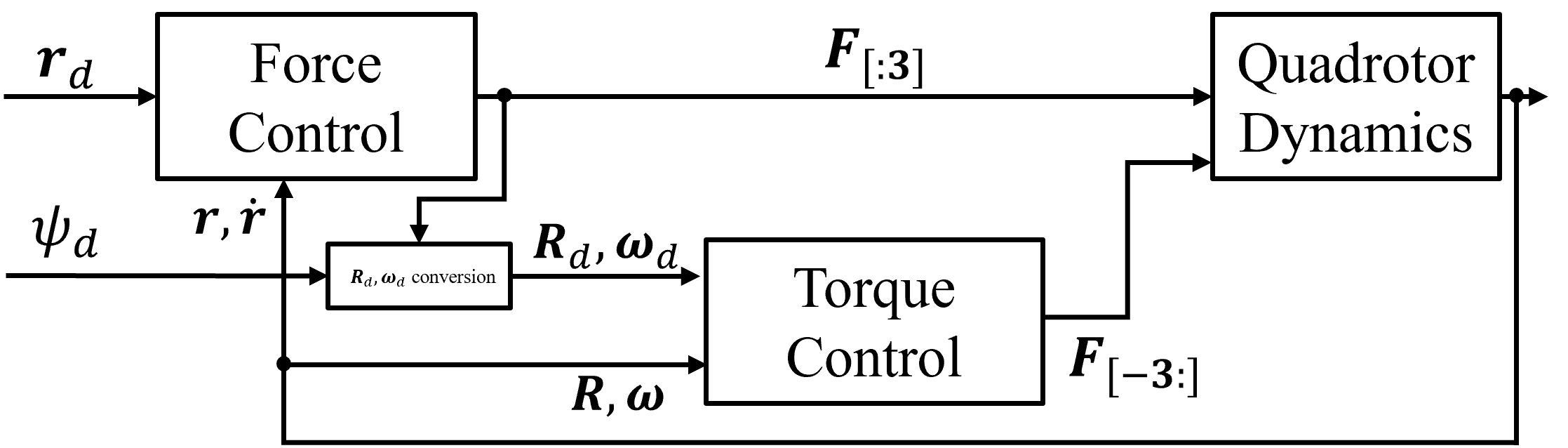}
\caption{Block diagram illustrating the cascaded control scheme.}
\label{cascaded_ctl}
\end{figure}

By applying the control law Eq.~\eqref{eq:int_ctl_law} to the quadrotor dynamics Eq.~\eqref{eq:model}, the error dynamics becomes:
\begin{equation}\label{eq:error_dyn}
    \bm{\ddot{\tilde{q}}} + 2\bm{\Lambda\dot{\tilde{q}}} + \bm{\Lambda^2\tilde{q}} = \bm{d} - \bm{\hat{d}}
\end{equation}
A combined error term inspired by the sliding mode approach is adopted as:
\begin{equation}\label{combined_e}
    \bm{s} = \bm{\dot{\tilde{q}}} + \bm{\Lambda\tilde{q}}
\end{equation}
Substituting for Eq.~\eqref{combined_e} and its first derivative into Eq~.\eqref{eq:error_dyn}, closed loop error dynamics becomes:
\begin{equation}
    \bm{\dot{s}} + \bm{\Lambda s} = \bm{d} - \bm{\hat{d}}
\end{equation}
It can be verified that, as $\bm{\hat{d}}\to \bm{d}$ for an ideal estimate, the combined error $\bm{s}$ and the tracking error $\bm{\tilde{q}}$ converge exponentially to zero. If not, the closed-loop dynamics is driven by the approximation error.

We then turn the control law Eq.~\eqref{eq:int_ctl_law} into an intelligent controller, by introducing an RBF network to estimate all uncertainty $\bm{d}$ with arbitrary precision $\bm{\varepsilon}$. This follows from the RBF network's universal approximation property.
\begin{equation}\label{eq:dhat}
    \hat{d}_i = \mathbf{w}^\top_i\bm{\varphi}_i(s_i)
\end{equation}
where $\hat{d}_i$ are the components of $\bm{\hat{d}}$, $ \mathbf{w}_i = [w_{i,1}, w_{i,2}, \dots , w_{i,n_i}]^\top$ is the weight vector and $\bm{\varphi}_i(s_i)=[\varphi_{i,1}, \varphi_{i,2}, \dots , \varphi_{i,n_i}]^\top$ is the vector with the activation functions $\varphi_{i,j} = \exp[-\frac{(s_i-c_{i,j})^2}{2\sigma^2_{i,j}}]$, with $i = 1 \dots 6$, $j = 1, 2, \dots, n_i$ and $n_i$ being the number of neurons in the hidden layer corresponding to $i$th component of $\bm{\hat{d}}$.
Suppose an ideal set of weights $\mathbf{w}^\ast$ that allows for an approximation of $\bm{d} = \mathbf{w}^{\ast\top}\bm{\varphi} + \bm{\varepsilon}$, the estimation error becomes $\tilde{\bm{d}}= \hat{\bm{d}}-\bm{d} = (\mathbf{w}-\mathbf{w}^\ast)^\top\bm{\varphi} - \bm{\varepsilon} = \tilde{\mathbf{w}}^\top\bm{\varphi} - \bm{\varepsilon}$. %This shows that the estimation error depends on how far the current weights are from the ideal.

To ensure boundedness and convergence conditions of the closed-loop system, a Lyapunov-based stability analysis is used as:
\begin{equation}\label{eq:lyapunov_can}
    V_i = \frac{1}{2}s^2_i + \frac{1}{2\eta_i}\tilde{\mathbf{w}}^\top_i\tilde{\mathbf{w}}_i
\end{equation}
whose time derivative becomes:
\begin{equation}\label{eq:der_lyapunov}
    \dot{V}_i = s_i\dot{s}_i + \eta^{-1}_i\mathbf{\tilde{w}}^\top_i\mathbf{\dot{w}}_i
\end{equation} since $\mathbf{w}^*$ is constant. After further respective substitutions, we obatain:
\begin{subequations}\label{eq:der_lyapunov1}
\begin{align}
    \dot{V}_i &= s_i(-\lambda_i s_i + d_i - \hat{d}_i) + \eta^{-1}_i\mathbf{\tilde{w}}^\top_i\mathbf{\dot{w}}_i \label{eq:der_lyapunov1a} \\
    \dot{V}_i &= s_i(-\lambda_i s_i - \tilde{\mathbf{w}}^\top_i\bm{\varphi}_i + \varepsilon_i) + \eta^{-1}_i\mathbf{\tilde{w}}^\top_i\mathbf{\dot{w}}_i \label{eq:der_lyapunov1b} \\
    \dot{V}_i &= -s_i(\lambda_i s_i - \varepsilon_i) + \eta^{-1}_i\mathbf{\tilde{w}}^\top_i(\mathbf{\dot{w}}_i - \eta_i s_i\bm{\varphi}_i) \label{eq:der_lyapunov1c}
\end{align}
\end{subequations}

Therefore, by updating $\mathbf{w}_i$ according to the learning rule $\mathbf{\dot{w}}_i = \eta\bm{s}_i\bm{\varphi}_i$, with $\eta$ as the learning rate. The time derivative $\dot{V}_i$ becomes:
\begin{equation}\label{eq:der_lyapunov4}
    \dot{V}_i = -s_i (\lambda_i s_i - \varepsilon_i)
\end{equation}
implying that $\dot{V}_i$ is negative definite when $|s_i| > \frac{\varepsilon_i}{\lambda_i}$ and the bounds of $\mathbf{w}_i$ cannot be ensured with $\mathbf{\dot{w}}_i = \eta\bm{s}_i\bm{\varphi}_i$ when $|s_i| \leq \frac{\varepsilon_i}{\lambda_i}$. To guaranty that $\mathbf{w}_i$ remains within a convex region $\mathcal{W} _i = \{ \mathbf{w}_i \in \mathbb{R}^n : \mathbf{w}_i^\top\mathbf{w}_i \leq \mu_i^2 \}$, we employ the projection algorithm \citep{Ioannou2006}, which is also used in~\citep{Da_Silva_Lima} for a trajectory tracking task:

\begin{equation}
    \mathbf{\dot{w}}_i = \begin{cases}
                            \eta_i s_i \bm{\varphi}_i,
                                        & \begin{aligned}
                                        &\text{if } \|\mathbf{w}_i\|_2 < \mu_i \\
                                        &\text{or } \|\mathbf{w}_i\|_2 = \mu_i
                                        \ \text{and } \eta_i s_i \mathbf{w}_i^\top\bm{\varphi}_i < 0
                                        \end{aligned}
                        \\[10pt]
                        \left(
                        \bm{I} - \dfrac{\mathbf{w}_i \mathbf{w}_i^\top}{\mathbf{w}_i^\top \mathbf{w}_i}
                        \right)
                        \eta_i s_i \bm{\varphi}_i,
                        & \text{otherwise}
                        \end{cases}
\end{equation}
where, $\mu_i$ is the desired upper bound of $\|\mathbf{w}_i\|_2$.

Since $\|\mathbf{w}_i(0)\|_2 \leq \mu_i$, it follows that $|s_i| \leq \frac{\varepsilon_i}{\lambda_i}$ and $\|\mathbf{w}_i(t)\|_2 \leq \mu_i$ as $t \to \infty$. From Eq.~\eqref{combined_e}, we could obtain:
\begin{equation} \label{ineq}
    -\lambda_i^{-1}\varepsilon_i \leq \dot{\tilde{q}}_i + \lambda_i\tilde{q}_i \leq \lambda_i^{-1}\varepsilon_i
\end{equation}

Multiplying Eq.~\eqref{ineq} by $e^{\lambda_i t}$:
\begin{subequations}\label{eq:ineq_exp}
\begin{align}
    -\lambda_i^{-1}\varepsilon_i e^{\lambda_i t} \leq (\dot{\tilde{q}}_i + \lambda_i\tilde{q}_i)e^{\lambda_i t} \leq \lambda_i^{-1}\varepsilon_i e^{\lambda_i t}\\
    -\lambda_i^{-1}\varepsilon_i e^{\lambda_i t} \leq \frac{d}{dt}(\tilde{q}_i e^{\lambda_i t}) \leq \lambda_i^{-1}\varepsilon_i e^{\lambda_i t}
\end{align}
\end{subequations}
and integrating the resulting inequality between $0$ and $t$, then dividing by $e^{\lambda_i t}$:
\begin{equation} \label{int_ineq}
    -\frac{\varepsilon_i}{\lambda_i^2}- \left[|\tilde{q}_i(0)| + \frac{\varepsilon_i}{\lambda_i^2} \right] e^{-\lambda_i t} \leq \tilde{q}_i \leq \frac{\varepsilon_i}{\lambda_i^2} + \left[|\tilde{q}_i(0)| + \frac{\varepsilon_i}{\lambda_i^2} \right] e^{-\lambda_i t}
\end{equation}
As $t \to \infty$, Eq.~\eqref{int_ineq} becomes:
\begin{equation} \label{int_ineq2}
    -\frac{\varepsilon_i}{\lambda_i^2} \leq \tilde{q}_i \leq \frac{\varepsilon_i}{\lambda_i^2}
\end{equation}
Applying Eq.~\eqref{int_ineq2} to Eq.~\eqref{ineq} yields:
\begin{equation} \label{int_ineq3}
    -2\frac{\varepsilon_i}{\lambda_i} \leq \dot{\tilde{q}}_i \leq 2\frac{\varepsilon_i}{\lambda_i}
\end{equation}
This implies that the proposed control scheme ensures the exponential convergence of the tracking error to the closed region $\mathcal{Q} = \{ (\tilde{\bm{q}}, \dot{\tilde{\bm{q}}}) \in \mathbb{R}^6 : |\tilde{q}_i| \leq \varepsilon_i\lambda_i^{-2}$ and $ |\dot{\tilde{q}}_i| \leq 2\varepsilon_i\lambda_i^{-1} $.

The choice of $\lambda_i$ the control gains that also affect the tracking error surfaces is intended to establish stable nominal convergence before adaptation is enabled. Large values of the same improve the system response, however, increased sensitivity to noise and oscillation is likely especially the position corresponding values, so these are chosen moderately. Tuning of the learning rate $\eta_i$ is done to balance adaptation speed and robustness: too small a value results in negligible benefit, while larger values produced noisier estimates and degraded tracking. The projection bounds $\mu_i$ are set to keep the weights of the neural network bounded, preventing drift while avoiding excessive clipping of the disturbance estimates. 

\section{EXPERIMENTATION} \label{experimentation}

The control scheme is evaluated numerically and experimentally to illustrate its effectiveness. Real flight and simulation experiments were done using Robot Operating System (ROS)2 Humble and Crazyflie 2.1 drone.

\subsection{Trajectory Tracking}

We employ the intelligent control scheme to track a spiral trajectory with reference position $\bm{r_d}(t) = [x_d(t), y_d(t),z_d(t)]$ and yaw orientation $\psi_d$. Position and velocity tracking errors are defined as the difference between the robot state and the desired state:
\begin{subequations}\label{eq:pose_error}
\begin{align}
    \bm{e_r} &= \bm{r} - \bm{r}_d \label{eq:pos_e}\\
    \bm{e_v} &= \bm{v} - \bm{v}_d \label{eq:orn_e}
\end{align}
\end{subequations}
The orientation and angular velocity error are expressed such that they evolve on the tangent space of the configuration manifold~\citep{Antal} as:
\begin{equation}\label{rotation_error}
    \bm{e_R} = 0.5(\bm{R}_d^\top \bm{R} - \bm{R}^\top \bm{R}_d)^\vee
\end{equation}
where, the $(.)^\vee$ operator is the inverse of the $(\hat{.}$) operator introduced in Eq.~\eqref{eq:omega}, and:
\begin{equation}
    \bm{e_\omega} = \bm{\omega} - \bm{R}^\top \bm{R_d} \bm{\omega_d}
\end{equation}

The dynamics of the quadrotor being differentially flat allows us to design reference trajectories in terms of position and yaw angle~\citep{Antal}. The reference rotation matrix used in Eq.\eqref{rotation_error} can then be expressed as $\bm{R}_d \in SO(3) = [\bm{r}_1, \bm{r}_2, \bm{r}_3]$ where
\begin{subequations}
    \begin{align}
        \bm{r}_3 &= \frac{\bm{F}_{[:3]}}{\|\bm{F}_{[:3]}\|} \\
        \bm{r}_2 &= \frac{\bm{r}_3 \times \bm{a}_\psi}{\|\bm{r}_3 \times \bm{a}_\psi\|} \\
        \bm{r}_1 &= \bm{r}_2 \times \bm{r}_3
    \end{align}
\end{subequations}
with $\bm{a}_\psi = [c\psi_d, s\psi_d, 0]^\top$. The vector $\bm{r}_3$ expresses the desired thrust direction, whereas $\bm{r}_1$ expresses the heading of the quadrotor, and $\bm{r}_2$ is always perpendicular to the other two vectors.

\begin{figure}[h!]
    \centering
    \begin{subfigure}{0.48\textwidth}
        \centering
        \includegraphics[width=0.95\linewidth]{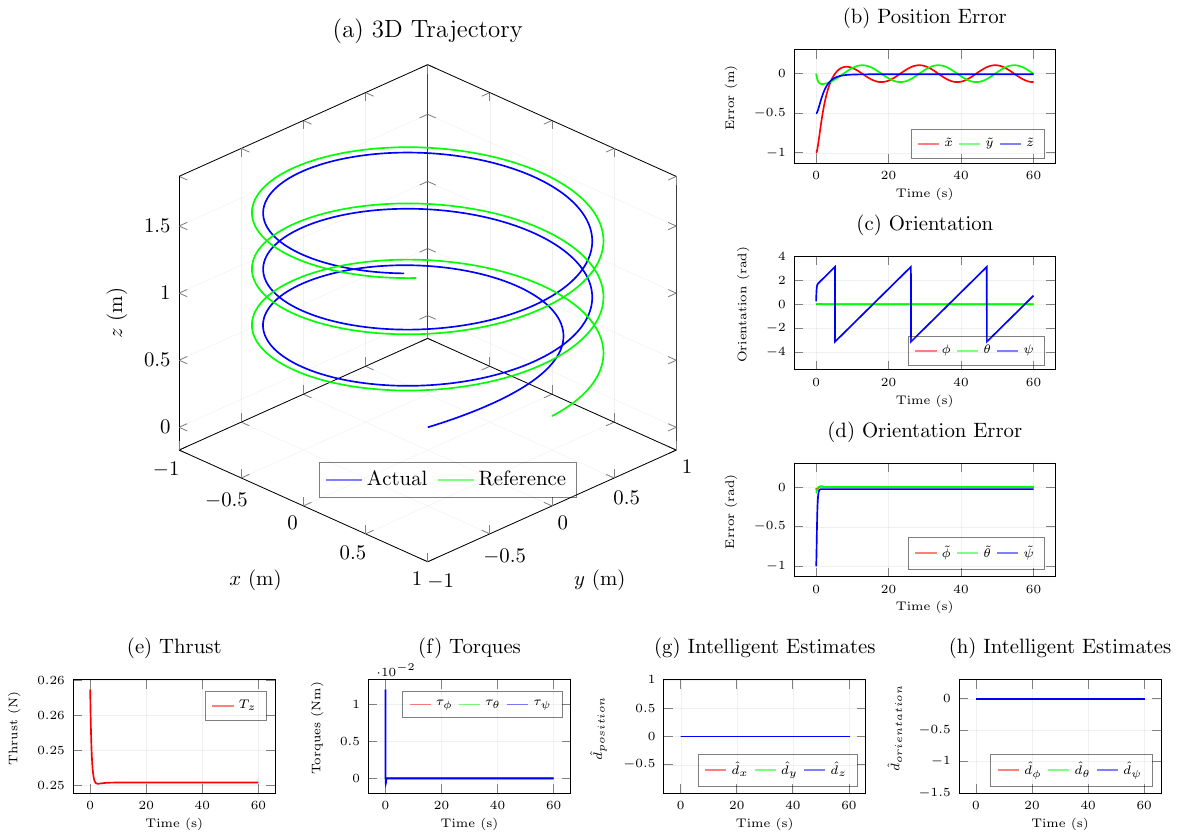}
        \caption{Feedback linearization}
        \label{fig:ns_fbl}
    \end{subfigure}
    \hfill
    \begin{subfigure}{0.48\textwidth}
        \centering
        \includegraphics[width=0.95\linewidth]{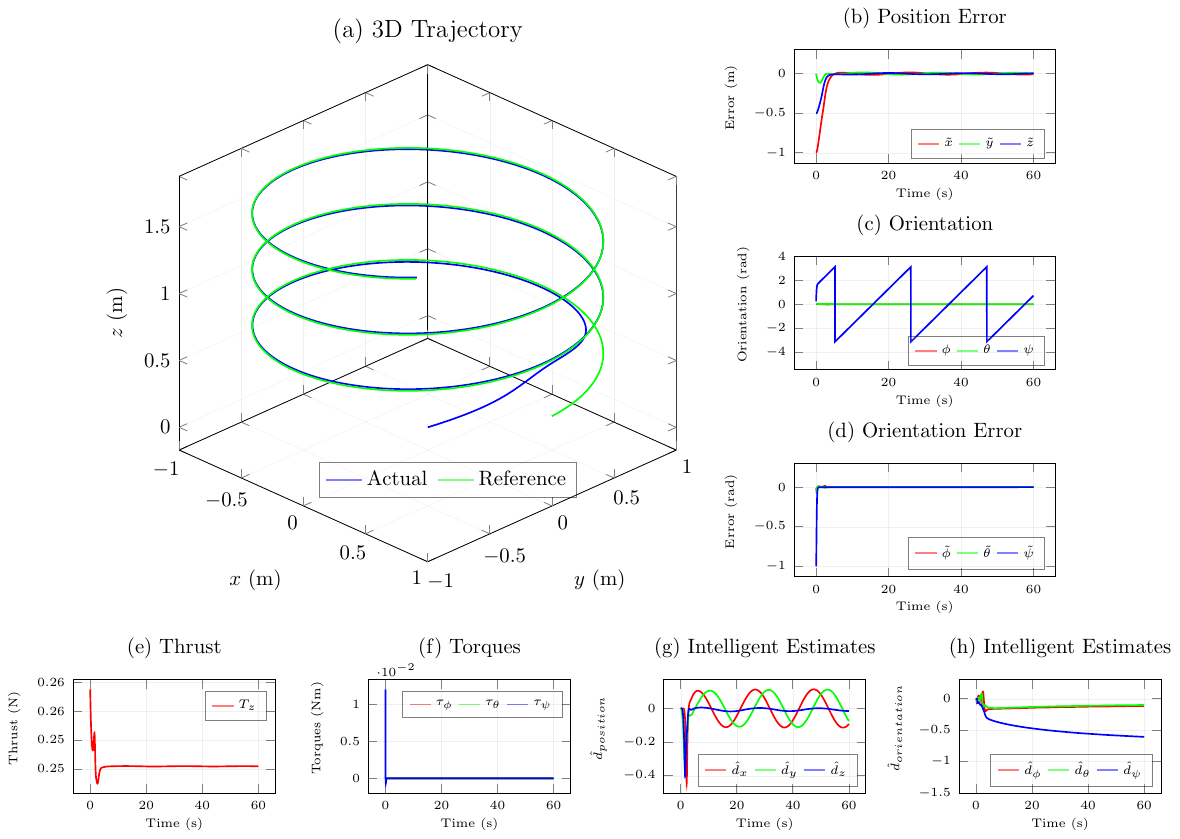}
        \caption{Intelligent controller}
        \label{fig:ns_int_ctl}
    \end{subfigure}
    \caption{Numeric simulation results of the controller tracking a spiral trajectory of 1.0 m radius with a step of 0.02 m/s at 0.3 rad/s angular speed.}
    \label{results_numeric}
\end{figure}

%\begin{figure}[h!]
%\centering
%\includegraphics[width=100mm]{Figures/NS_int_ctl.pdf}
%\caption{Numeric simulation results of the intelligent controller tracking a spiral trajectory of 1.0 m radius with a step of 0.02 m/s at 0.3 rad/s angular speed.}
%\label{results_numeric}
%\end{figure}

\subsection{Numerical Simulations} \label{numerical}

Numerical simulations were performed in python by applying the intelligent control scheme Eq.~\eqref{eq:int_ctl_law} to the quadrotor dynamics. The fourth-order Runge-Kutta was employed to solve ordinary differential equations. A 400 Hz sampling rate was used for both the controller and the dynamics model. The quadrotor parameters were set as: $m = 0.025$ kg, $\bm{J} = [16.5717\times10^{-6}, 16.5717\times10^{-6}, 29.2616\times10^{-6}]$ and $g = 9.81$ m/s\textsuperscript{2}. The control parameters were set as $\lambda_i = 1$ and a learning rate of $\eta_i = 1$ for the position sub-space $i = 1,2,3$ while $\bm{\Lambda}_{[-3:]} = [10, 10, 20]$ and $\eta_i = 1$ for the orientation sub-space $i = 4,5,6$. A single layer of the RBF network is considered with input $s_i$ and output $\hat{d}_i$ for each state variable. Six neurons are set for the single hidden layer whose centers $\bm{c}_i = \begin{bmatrix}-0.1 &-0.05 &-0.025 &0.025 &0.05 &0.1\end{bmatrix}$ and widths $\bm{\sigma}_i = \begin{bmatrix}0.1 &0.08 &0.06 &0.06 &0.08 &0.1\end{bmatrix}$ are chosen to balance a trade-off between control precision and computational burden. The initialization of the weights $\mathbf{w}_i = 0$ is chosen to allow the robot learn from its interaction with the environment. To evaluate the performance of the intelliegent control scheme in dealing with uncertainities, unmodeled external disturbances were introduced as $\bm{d} = -\bm{\zeta\dot{q}}$, where $\bm{\zeta} = \text{diag}(\frac{k_x}{m}, \frac{k_y}{m}, \frac{k_z}{m}, \frac{k_\phi}{I_{xx}}, \frac{k_\theta}{I_{yy}}, \frac{k_\psi}{I_{zz}})$, $k_{(\cdot)} = 0.01$ for linear drag coeeficient, and $0.001$ for angular drag coeeficient. Figure~\ref{results_numeric} shows the trajectory tracking performance for a spiral trajectory defined by $r = 1.0$ m, $\omega = 0.3$ rad/s  and a step in the $z$-direction of $0.02$ m. Table~\ref{error_comp_num} shows a comparison in terms of the steady-state root mean square error (RMSE) in position, yaw orientation and angular velocity for both the intelligent controller and the pure feedback linearization controller.

\begin{table}[h]
\centering
\caption{A comparative analysis of the controller trajectory tracking performance in terms of Steady-State RMSE for the numeric simulation.}
\label{error_comp_num}
\begin{tabular}{lccccc}
\hline
Setting & RMSE$_x$ [m] & RMSE$_y$ [m] & RMSE$_z$ [m] & RMSE$_\psi$ [rad] & RMSE$_\omega$ [rad/s] \\
\hline
FBL$_{\eta_i=0}$    &  \(0.0738\)  & \(0.0764\)  & \(0.0082\)  & \(0.0026\)  & \(0.0001\) \\
IC$_{\eta_i=1.0}$  & \(0.0098\) & \(0.0086\) & \(0.0056\) & \(0.0012\) & \(0.0001\) \\
\hline
\end{tabular}
\end{table}

\subsection{Gazebo Simulation and Real Experiments}

The performance of the intelligent control scheme is further evaluated in ROS2 Humble with Crazyflie drone in Gazebo Harmonic simulation and on real hardware making use of its CrazySwarm API. Gazebo physics and the Unified Robot Description Format (URDF) were used to model the quadrotor dynamics and environment interaction in simulation. The quadrotor state estimation in simulation relied on odometry from Gazebo, whereas odometry from onboard IMU was used for the real drone. 
%Because the crazyflie API requires twist velocity commands, we compute the required velocities from the controller output. The onboard low-level controller on the drone converts the linear and angular velocities into motor speeds. 
The quadrotor parameters were set as in the numerical simulation subsection~\ref{numerical}.

\begin{figure}[h!]
    \centering
    \begin{subfigure}{0.48\textwidth}
        \centering
        \includegraphics[width=0.95\linewidth]{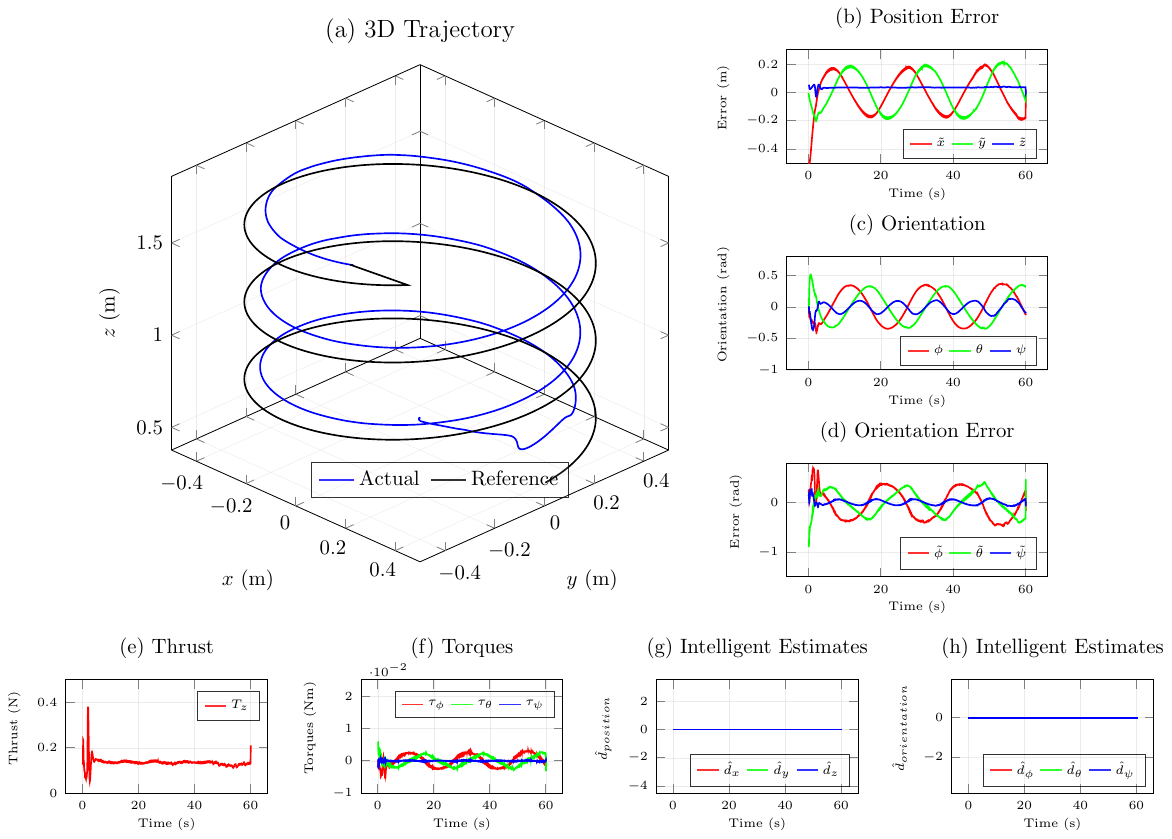}
        \caption{Feedback linearization}
        \label{results_fbl}
    \end{subfigure}
    \hfill
    \begin{subfigure}{0.48\textwidth}
        \centering
        \includegraphics[width=0.95\linewidth]{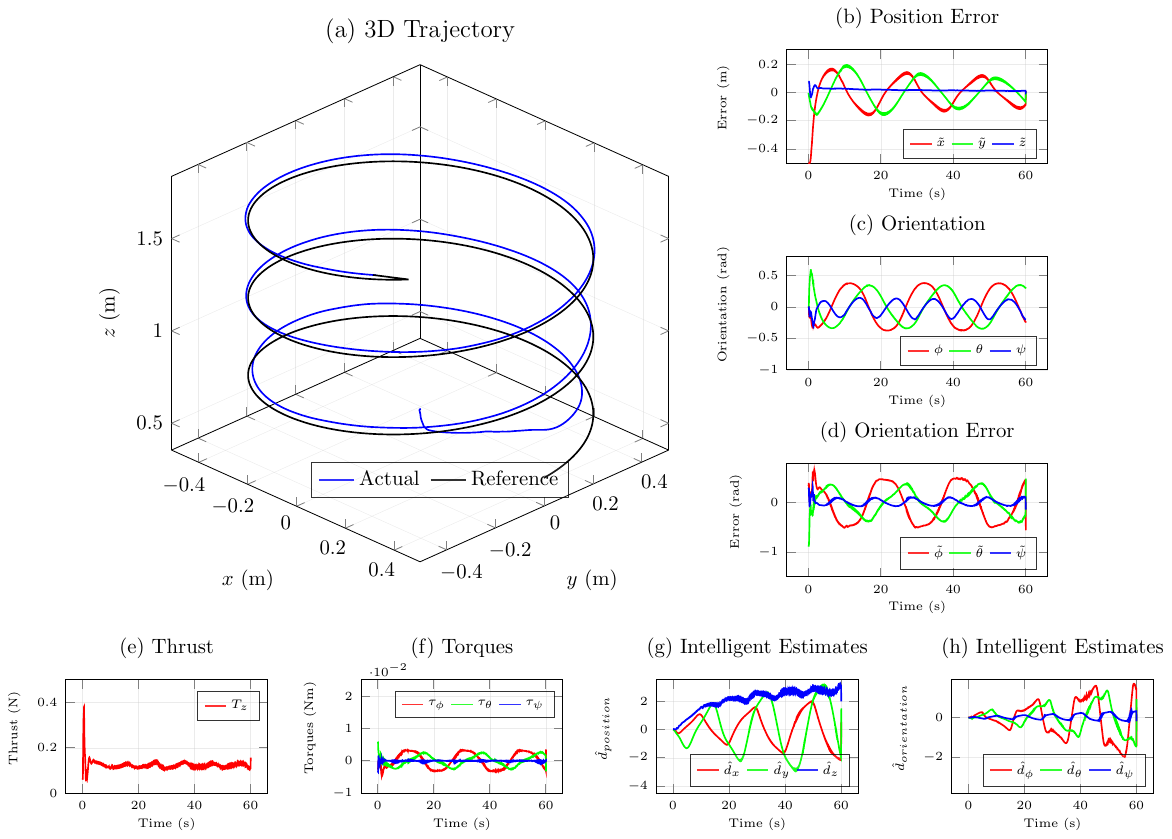}
        \caption{Intelligent controller (Case 2)}
        \label{results_intctl}
    \end{subfigure}
    \caption{Gazebo simulation results of the controller on Crazyflie quadrotor tracking a spiral trajectory of 0.5 m radius with a step of 0.02 m/s at 0.3 rad/s angular speed.}
    \label{results_gz}
\end{figure}

Gazebo simulation control parameters were set as $\bm{\lambda}_{1:3} = [5.0, 5.0, 10.0]$ for the position subspace, while $\bm{\lambda}_{4:6} = [20, 20, 20]$ for the orientation subspace. A single layer of the RBF network is considered with input $s_i$ and output $\hat{d}_i$ for each state variable with projection-bounded weights. The centers are chosen to span the range of sliding error as $\bm{c}_{xy} = [-1.0, -0.5, -0.25, 0.25, 0.5, 0.5, 1.0]$, $\bm{c}_{z} = [0.25, 0.3, 0.35, 0.4, 0.45, 0.5]$, $\bm{c}_{\psi} = [-2.0, -0.5 -0.25, 0.25, 0.5, 2.0]$, $\bm{c}_\omega = [-7.5, -4.0, -0.5, 0.5, 4.0, 7.5]$ with the corresponding widths chosen based on the distribution of the centers to balance performance. 
A limitation of this simulator setup is that roll and pitch rate commands are not directly applied and no onboard stabilization is in effect as observed in the plots. This underactuation prevents full attitude tracking. To evaluate the impact of the intelligent control scheme, we set the learning rate of the RBF network as $\eta_i = 0$ leading to a pure feedback linearization controller, figure~\ref{results_fbl}. Figure~\ref{results_intctl} shows the trajectory tracking results of the intelligent control scheme. It can be observed that, without prior modeling of air drag and actuator dynamics, the controller is capable of tracking the reference trajectory with reduced position error. Table~\ref{error_comp} shows a comparison in terms of steady-state root mean square error in position, yaw orientation, and angular velocity for both the intelligent controller and the baseline feedback linearization controller. The results compare three controller settings indicating stable steady-state tracking across five runs per setting and the effect of the Gaussian RBF networks when used to compensate for unmodeled system disturbances. In both case 1 and case 2, we observe an improved translational tracking relative to baseline in all position states, with case 2 producing a mean position-norm RMSE improvement of about $34.8\%$. There exists a clear trade-off, improving translational performance increases rotational aggressiveness.

\begin{table}[h]
\centering
\caption{A comparative analysis of the controller trajectory tracking performance in terms of Steady-State RMSE computed on $20$--$60~\mathrm{s}$ for the Gazebo simulation. (Case 1: $\eta_i = 0.01$, $W_i \leq 0.3$; Case 2: $\eta_{x,z} = 0.12, \eta_{y} = 0.2, \eta_{\phi\theta\psi} = 0.01$, $W_i \leq 2.0$)}
\label{error_comp}
\begin{tabular}{lccccc}
\hline
Setting & RMSE$_x$ [m] & RMSE$_y$ [m] & RMSE$_z$ [m] & RMSE$_\psi$ [rad] & RMSE$_\omega$ [rad/s]   \\
\hline
FBL$_{\eta_i=0}$ & \(0.1193 \pm 0.0023\)   & \(0.1331 \pm 0.0028\) & \(0.0344 \pm 0.0006\) & \(\bm{0.0433} \pm 0.0018\) & \(\bm{0.0379} \pm 0.0033\)\\
IC (Case 1) & \(0.1144 \pm 0.0004\) & \(0.1286 \pm 0.0003\) & \(0.0285 \pm 0.0001\) & \(0.0442 \pm 0.0003\) & \(0.0383 \pm 0.0004\)\\
IC (Case 2) & \(\bm{0.0828} \pm 0.0004\) & \(\bm{0.0838} \pm 0.0004\) & \(\bm{0.0138} \pm 0.0001\) & \(0.0602 \pm 0.0004\) & \(0.0567 \pm 0.0011\)\\
\hline
\end{tabular}
\end{table}

\begin{figure}[h!]
    \centering
    \begin{subfigure}{0.48\textwidth}
        \centering
        \includegraphics[width=0.95\linewidth]{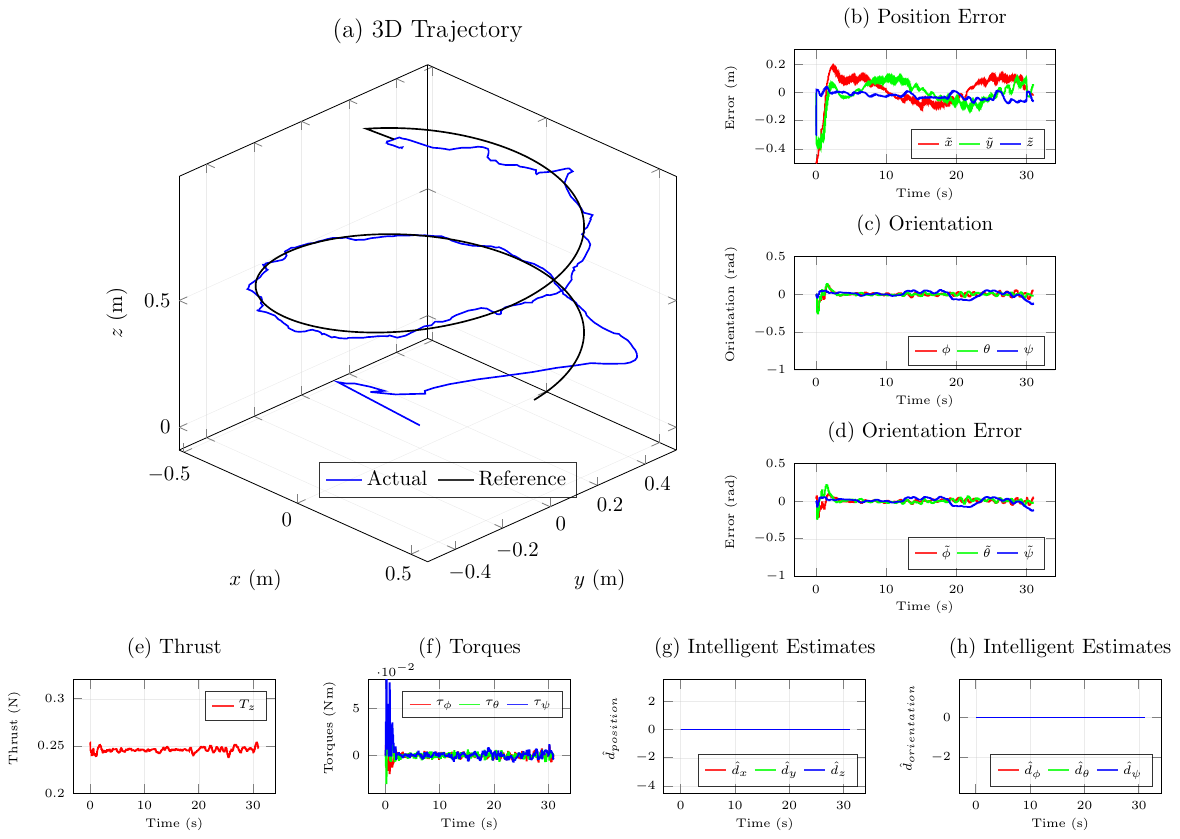}
        \caption{Feedback linearization}
        \label{cf_fbl}
    \end{subfigure}
    \hfill
    \begin{subfigure}{0.48\textwidth}
        \centering
        \includegraphics[width=0.95\linewidth]{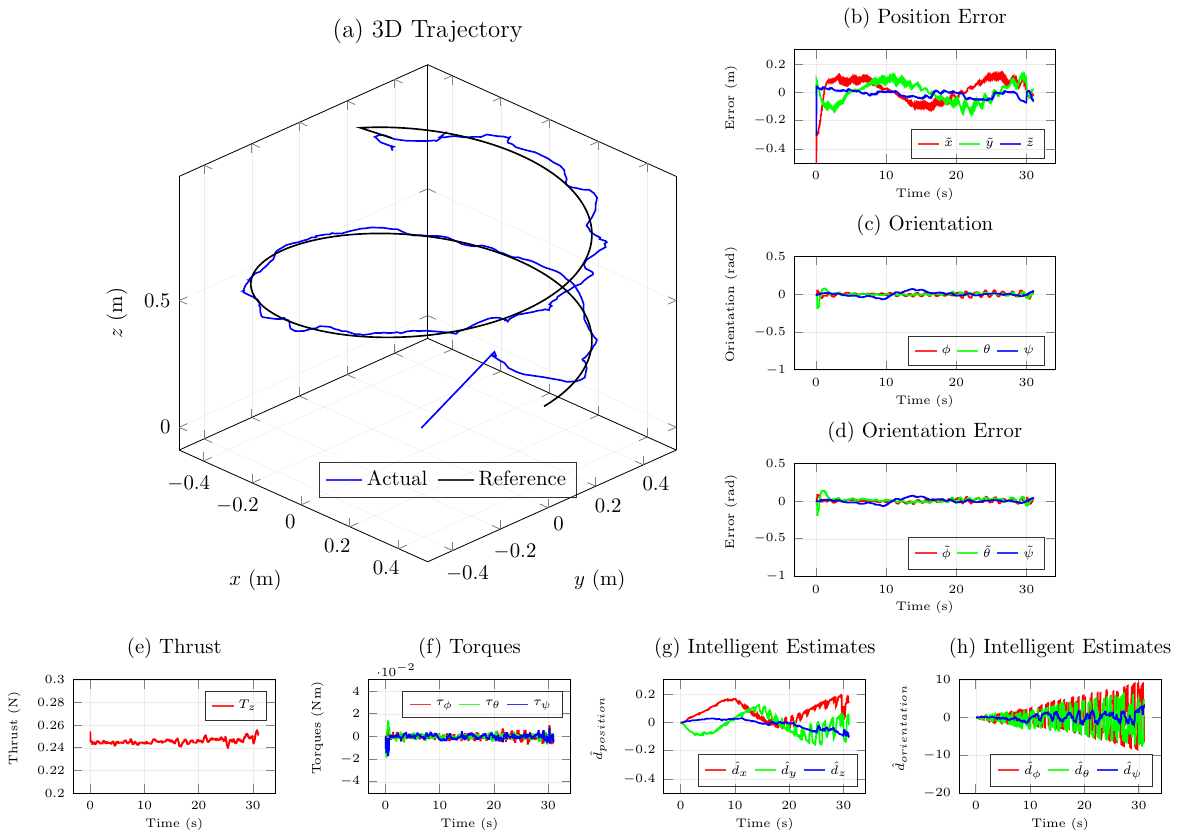}
        \caption{Intelligent controller}
        \label{cf_int}
    \end{subfigure}
    \caption{Results of the controller on Real flight test of the Crazyflie quadrotor tracking a spiral trajectory of 0.5 m radius with a step of 0.02 m/s at 0.3 rad/s angular speed.}
    \label{results_real}
\end{figure}

The real crazyflie tests utilized control parameters set as $\lambda_i = 1$ for the position sub-space $i = 1,2,3$ while $\bm{\Lambda}_{[-3:]} = [10, 10, 15]$ for the orientation sub-space $i = 4,5,6$. The centers of the hidden layer chosen to span the sliding error are set as $\bm{c}_{xyz} = \begin{bmatrix}-0.2 & -0.1 &-0.05 &0.05 &0.1 &0.2\end{bmatrix}$, $\bm{c}_{\phi\theta\psi} = \begin{bmatrix}-10.0 & -5.0 & -2.5 & 2.5 & 5.0 & 10.0\end{bmatrix}$, while the widths are determined based on the distribution of the centers. Figure~\ref{results_real} shows the trajectory tracking performance of both the feedback linearization and the intelligent controller. 
Table~\ref{error_comp_real} shows a comparison of steady state RMSE for different controller settings. The results indicate stable steady-state across five runs per setting with the intelligent controller (case 3) showing improved translational tracking relative to the baseline. A mean RMSE improvement of the position-norm of about $7.13\%$, while $49.27\%$ in yaw orientation is achieved with roll and pitch stabilization.

\begin{table}[h]
\centering
\caption{A comparative analysis of the controller trajectory tracking performance in terms of Steady-State RMSE computed on $5$--$30~\mathrm{s}$ for the real-flight experiments. (Case 3: $\eta_{x,z} = 0.08, \eta_{y} = 0.1, \eta_{\phi\theta\psi} = 0.1$)}
\label{error_comp_real}
\begin{tabular}{lcccccc}
\hline
Setting & RMSE$_x$ [m] & RMSE$_y$ [m] & RMSE$_z$ [m] & RMSE$_\psi$ [rad] & RMSE$_\omega$ [rad/s] \\
\hline
FBL$_{\eta_i=0}$ & \(0.0745 \pm 0.0028\) & \(0.0694 \pm 0.0056\) & \(0.0392 \pm 0.0152\) & \(0.0376 \pm 0.0127\) & \(3.2704 \pm 1.1655\) \\
IC$_{\eta_i=case 3}$ & \(\bm{0.0725} \pm 0.0033 \) & \(\bm{0.0664} \pm 0.0028 \) & \(\bm{0.0262} \pm 0.0068 \) & \(\bm{0.0191} \pm 0.0069 \) & \(\bm{1.7313} \pm 0.3833 \) \\
%IC$_{\eta_i=0.01}$ & 0.0768 $\pm$ 0.0025 & 0.0713 $\pm$ 0.0028 & 0.0486 $\pm$ 0.0117 & 0.0244 $\pm$ 0.0079 & 2.2348 $\pm$ 1.5190 \\
%IC$_{\eta_i=0.08}$& 0.0748 $\pm$ 0.0068 & 0.0696 $\pm$ 0.0061 & 0.0456 $\pm$ 0.0276 & 0.0250 $\pm$ 0.0087 & 2.8637 $\pm$ 1.2288 \\
%IC$_{\eta_i=0.1}$  & 0.0762 $\pm$ 0.0054 & 0.0679 $\pm$ 0.0031 & 0.0485 $\pm$ 0.0273 & 0.0168 $\pm$ 0.0016 & 1.5879 $\pm$ 0.0388 \\
%IC$_{\eta_i=1.0}$ & 0.0792 $\pm$ 0.0046 & 0.0728 $\pm$ 0.0016 & 0.0599 $\pm$ 0.0186 & 0.0153 $\pm$ 0.0028 & 1.5990 $\pm$ 0.2554 \\
\hline
\end{tabular}
\end{table}

\section{CONCLUSION} \label{conclusion}
This paper presents a machine learning-based feedback linearization controller for quadrotor operating under uncertain dynamics. An online Gaussian RBF network was developed to estimate unmodeled dynamics and unknown nonlinearities in the control system. A Lyapunov-based controller design was utilized herein ensuring closed-loop stability and asymptotic convergence of trajectory tracking errors. Simulation and real-flight experiments with the Crazyflie drone confirmed that the proposed approach achieves trajectory tracking with error convergence and robustness to unmodeled disturbances.
Future extensions will incorporate the use of reinforcement learning-based parameter tuning and adaptation, extension of the framework to multi-robot coordination, and integration of obstacle-aware navigation.

\section{ACKNOWLEDGEMENTS}
“This project has received funding from the European Union’s Horizon Europe research and innovation programme under the Marie Skłodowska-Curie Actions grant agreement No. 101125250."

“Funded by the European Union. Views and opinions expressed are however those of the author(s) only and do not necessarily reflect those of the European Union or European Research Executive Agency (REA). Neither the European Union nor the granting authority can be held responsible for them.”

"We acknowledge the financial support of the Finnish Ministry of Education and Culture through the Intelligent Work Machines Doctoral Education Pilot Program (IWM VN/3137/2024-OKM-4)."

"We acknowledge the use of the TIERS lab drone arena  through the support of the Academy of Finland’s AeroPolis project (Grant No. 348480)."

\section{REFERENCES} 

\bibliographystyle{eurodiname2026}
\renewcommand{\refname}{}
\bibliography{bibfile}

\section{RESPONSIBILITY NOTICE}

The authors are solely responsible for the printed material included in this paper.

\section{APPENDIX} \label{appendix}

\subsection{Quadrotor Lagrangian Dynamics}

%\subsubsection{Angular velocity (body frame)}
%\[\bm{\omega}_B = \left[\begin{matrix}\dot{\phi} - \dot{\psi} S{\left(\theta \right)}\\\dot{\psi} S{\left(\phi \right)} C{\left(\theta \right)} + \dot{\theta} C{\left(\phi \right)}\\\dot{\psi} C{\left(\phi \right)} C{\left(\theta \right)} - \dot{\theta} S{\left(\phi \right)}\end{matrix}\right] \]

\subsubsection{Mass matrix}
\[\displaystyle \bm{M(q)} = \left[\begin{matrix}m & 0 & 0 & 0 & 0 & 0\\0 & m & 0 & 0 & 0 & 0\\0 & 0 & m & 0 & 0 & 0\\0 & 0 & 0 & I_{x} & 0 & - I_{x} S{\left(\theta \right)}\\0 & 0 & 0 & 0 & I_{y} C^{2}{\left(\phi \right)} + I_{z} S^{2}{\left(\phi \right)} & \frac{\left(I_{y} - I_{z}\right) \left(S{\left(2 \phi - \theta \right)} + S{\left(2 \phi + \theta \right)}\right)}{4}\\0 & 0 & 0 & - I_{x} S{\left(\theta \right)} & \frac{\left(I_{y} - I_{z}\right) \left(S{\left(2 \phi - \theta \right)} + S{\left(2 \phi + \theta \right)}\right)}{4} & I_{x} S^{2}{\left(\theta \right)} + I_{y} S^{2}{\left(\phi \right)} C^{2}{\left(\theta \right)} + I_{z} C^{2}{\left(\phi \right)} C^{2}{\left(\theta \right)}\end{matrix}\right] \]

\subsubsection{Coriolis/Centrifugal matrix}
\[\displaystyle
    \bm{C(q, \dot{q})} = \begin{bmatrix}0 &0 & 0 &0 &0 &0 \\0 &0 & 0 &0 &0 &0 \\0 &0 & 0 &0 &0 &0 \\0 &0 & 0 &0 &c_{3,4} &c_{3,5} \\0 &0 & 0 &c_{4,3} &c_{4,4} &c_{4,5} \\0 &0 & 0 &c_{5,3} &c_{5,4} &c_{5,5}\end{bmatrix}
\]
where,

$c_{3,4} = - \frac{\dot{\psi} \left(2 I_{x} C{\left(\theta \right)} + \left(I_{y} - I_{z}\right) \left(C{\left(2 \phi - \theta \right)} + C{\left(2 \phi + \theta \right)}\right)\right)}{4} + \frac{\dot{\theta} \left(I_{y} - I_{z}\right) S{\left(2 \phi \right)}}{2}$

$c_{3,5} = - \dot{\psi} \left(I_{y} - I_{z}\right) S{\left(\phi \right)} C{\left(\phi \right)} C^{2}{\left(\theta \right)} - \frac{\dot{\theta} \left(2 I_{x} C{\left(\theta \right)} + \left(I_{y} - I_{z}\right) \left(C{\left(2 \phi - \theta \right)} + C{\left(2 \phi + \theta \right)}\right)\right)}{4}$

$c_{4,3} = \frac{\dot{\psi} \left(2 I_{x} C{\left(\theta \right)} + \left(I_{y} - I_{z}\right) \left(C{\left(2 \phi - \theta \right)} + C{\left(2 \phi + \theta \right)}\right)\right)}{4} - \frac{\dot{\theta} \left(I_{y} - I_{z}\right) S{\left(2 \phi \right)}}{2}$

$c_{4,4} = \frac{\dot{\phi} \left(- I_{y} + I_{z}\right) S{\left(2 \phi \right)}}{2}$

$c_{4,5} = \frac{\dot{\phi} \left(2 I_{x} C{\left(\theta \right)} + \left(I_{y} - I_{z}\right) \left(C{\left(2 \phi - \theta \right)} + C{\left(2 \phi + \theta \right)}\right)\right)}{4} + \dot{\psi} \left(- I_{x} + I_{y} S^{2}{\left(\phi \right)} + I_{z} C^{2}{\left(\phi \right)}\right) S{\left(\theta \right)} C{\left(\theta \right)}$

$c_{5,3} = \dot{\psi} \left(I_{y} - I_{z}\right) S{\left(\phi \right)} C{\left(\phi \right)} C^{2}{\left(\theta \right)} - \frac{\dot{\theta} \left(2 I_{x} C{\left(\theta \right)} + \left(- I_{y} + I_{z}\right) \left(C{\left(2 \phi - \theta \right)} + C{\left(2 \phi + \theta \right)}\right)\right)}{4}$

$c_{5,4} = - \frac{\dot{\phi} \left(2 I_{x} C{\left(\theta \right)} - \left(I_{y} - I_{z}\right) \left(C{\left(2 \phi - \theta \right)} + C{\left(2 \phi + \theta \right)}\right)\right)}{4} - \dot{\psi} \left(- I_{x} + I_{y} S^{2}{\left(\phi \right)} + I_{z} C^{2}{\left(\phi \right)}\right) S{\left(\theta \right)} C{\left(\theta \right)} - \frac{\dot{\theta} \left(I_{y} - I_{z}\right) \left(C{\left(2 \phi - \theta \right)} - C{\left(2 \phi + \theta \right)}\right)}{4}$

$c_{5,5} = (\frac{\dot{\phi} (I_{y} - I_{z}) (S{\left(2 \phi - \theta \right)} + S{(2 \phi + \theta)})}{4}) - \dot{\theta} (- I_{x} + I_{y} S^{2}{(\phi)} + I_{z} C^{2}{(\phi)}) S{(\theta)}) C(\theta)$

%\subsubsection{Gravity vector}
%\[\displaystyle \bm{G(q)} = \left[\begin{matrix}0\\0\\g m\\0\\0\\0\end{matrix}\right] \]

\end{document}